%% file: MapGCLR.tex
\documentclass[letterpaper, 10 pt, conference]{ieeeconf}
\IEEEoverridecommandlockouts    % to override locked commands
\usepackage{multirow}
\usepackage{lipsum}
\usepackage{amsmath}
\usepackage{amssymb}
\usepackage[ruled,vlined]{algorithm2e}
\usepackage{graphicx}
\usepackage{adjustbox}
\usepackage{svg}
\usepackage{siunitx}
\usepackage{pifont}
\usepackage{xspace}
\usepackage[dvipsnames]{xcolor}
\graphicspath{{./Figures/}}

\usepackage{todonotes}
\usepackage[nolist]{acro}
\usepackage{multirow}
\usepackage{pgfplots}
\pgfplotsset{compat=1.18}

\usepackage[font=small]{caption}
\usepackage[T1]{fontenc}
\usepackage[utf8]{inputenc}

\usepackage{balance}

\usepackage{hyperref}
\hypersetup{
    colorlinks=true,
    linkcolor=black,
    citecolor=black,
    filecolor=black,
    urlcolor=black,
    pdfauthor={Anonymous}
}
\usepackage[capitalise]{cleveref}

\newlength{\IEEEleft}
\usepackage{balance}
\usepackage[style=ieee,
            doi=false,
            url=false,
            mincitenames=1,
            maxcitenames=1,
            minbibnames=6,
            maxbibnames=6,
            backend=biber]{biblatex}  % supports bibliography with BibLaTeX
\newcommand{\eg}{e.\,g.\xspace}     % for example
\title{\LARGE \bf MapGCLR: Geospatial Contrastive Learning of Representations\\for Online Vectorized HD Map Construction}

\author{
    Jonas Merkert$^{1*}$,
    Alexander Blumberg$^{1*}$,
    Jan-Hendrik Pauls$^{1}$,
    and 
    Christoph Stiller$^{1}$% <-this % stops a space
    \thanks{$^{*}$These authors contributed equally to this work.}
    \thanks{
        $^{1}$Institute of Measurement and Control Systems, Karlsruhe Institute of Technology (KIT),
        Karlsruhe, Germany
        {\tt\footnotesize \{alexander.blumberg, jonas.merkert, jan-hendrik.pauls, christoph.stiller\}@kit.edu}
    }%
}

\input{sections/abbreaviations}

\usepackage{textcomp}
\usepackage[absolute,showboxes]{textpos}

\newcommand\copyrighttext{%
  \footnotesize \copyright 2026 IEEE.  Personal use of this material is permitted.  Permission from IEEE must be obtained for all other uses, in any current or future media, including reprinting/republishing this material for advertising or promotional purposes, creating new collective works, for resale or redistribution to servers or lists, or reuse of any copyrighted component of this work in other works.}

\newcommand\copyrightnotice{%
\begin{tikzpicture}[remember picture,overlay]
\node[anchor=south,yshift=10pt] at (current page.south) {\fbox{\parbox{\dimexpr\textwidth-\fboxsep-\fboxrule\relax}{\copyrighttext}}};
\end{tikzpicture}%
}

\begin{document}
	
	\maketitle
	\copyrightnotice

    \thispagestyle{empty}
	\pagestyle{empty}

    \input{sections/todo}
    \input{sections/00_abstract}
    \input{sections/01_intro}
    \input{sections/02_related_work}
    \input{sections/03_methodology}
    \input{sections/04_evaluation}
    \input{sections/05_conclusion}
    \input{sections/06_acknowledgement}

	%%%%%%%%%%%%%%%%%%%%%%%%%%%%%%%%%%%%%%%%%%%%%%%%%%%%%%%%%%%%%%%%%%
	% %\addtolength{\textheight}{-12cm}
	% %\vspace{10mm}
	% \bibliographystyle{IEEEtran}
	% % Your .bib file here
	% \bibliography{references}
    
    \balance % balance the columns of the bibliography
    \printbibliography

\end{document}

%% file: sections/abbreaviations.tex
\DeclareAcronym{hd}{
  short = HD,
  long  = high-definition,
}

\DeclareAcronym{sd}{
  short = SD,
  long  = standard-definition,
}

\DeclareAcronym{iou}{
  short = IoU,
  long  = Intersection over Union,
}

\DeclareAcronym{bev}{
  short = BEV,
  long  = birds-eye-view,
}

\DeclareAcronym{gru}{
  short = GRU,
  long = gated recurrent unit
}

\DeclareAcronym{ssl}{
  short = SSL,
  long = self-supervised learning
}

\DeclareAcronym{pca}{
  short = PCA,
  long = principal component analysis
}

%% file: sections/todo.tex
% \section*{TODO}
%\todo[inline]{Abstract schreiben}
%\todo[inline]{Einleitung schreiben}
%\todo[inline]{Related Work schreiben}
%\todo[inline]{Evaluation schreiben}
% \todo[inline]{Conclusion schreiben}

%% file: sections/00_abstract.tex
\begin{abstract}
    Autonomous vehicles rely on map information to understand the world around them.
	However, the creation and maintenance of offline \ac{hd} maps remains costly. A more scalable alternative lies in online \ac{hd} map construction, which only requires map annotations at training time.
    To further reduce the need for annotating vast training labels, self-supervised training provides an alternative.
    
    This work focuses on improving the latent \ac{bev} feature grid representation within a vectorized online \ac{hd} map construction model by enforcing geospatial consistency between overlapping \ac{bev} feature grids as part of a contrastive loss function.
    To ensure geospatial overlap for contrastive pairs, we introduce an approach to analyze the overlap between traversals within a given dataset and generate subsidiary dataset splits following adjustable multi-traversal requirements.
    We train the same model supervised using a reduced set of single-traversal labeled data and self-supervised on a broader unlabeled set of data following our multi-traversal requirements, effectively implementing a semi-supervised approach.
    Our approach outperforms the supervised baseline across the board, both quantitatively in terms of the downstream tasks vectorized map perception performance and qualitatively in terms of segmentation in the \ac{pca} visualization of the \ac{bev} feature space.
\end{abstract}

\begin{keywords}
Online \ac{hd} Map Construction, Automated Driving, Semi-Supervised Learning, Contrastive Learning 
\end{keywords}

%% file: sections/01_intro.tex
\acresetall

\section{Introduction}
\label{sec:introduction}

\Ac{hd} maps remain highly relevant as foundation for the planning task in current software stacks for autonomous vehicles~\cite{Kato2018Autoware}. Compared to \ac{sd} maps, \ac{hd} maps provide more detailed, precise, and semantic information, including, but not limited to, the actual geometry of road elements, such as lane dividers or road boundaries, as well as their relations~\cite{Lanelet2}.

The creation and constant maintenance of such \ac{hd} maps remain resource-intensive, requiring regular mapping using mobile mapping platforms equipped with highly precise sensors and reference global localization systems, as well as at least partial manual annotations.
\input{sections/figures/overlap}

Hence, recent research has focus on constructing \ac{hd} maps while driving.
These online methods predict vectorized \ac{hd} map-like representations of the local surrounding in real-time based on 360° vision input~\autocite{liao_MapTR, liao2024maptrv2, chen_maptracker_2025}, as well as possible prior knowledge~\autocite{immel_m3tr,immel_SDTagNet2025}.
Although this dramatically reduces the necessity for valid \ac{hd} maps on a global scale, these learned approaches still rely on a well-distributed set of training data and are susceptible to missing corner cases.
Perspectively, providing such training data remains the major bottleneck for scalable online \ac{hd} map construction.

This work strikes to overcome this bottleneck by focusing on improving the intermediate \ac{bev} feature grid representation present in vectorized \ac{hd} map construction models. Using self-supervised contrastive learning, we enforce a consistent latent space between geospatially overlapping \ac{bev} feature grids.

The main contributions of this work are:
\begin{itemize}
    \item As necessary foundation for geospatial consistency, we propose an approach to analyze the geospatial overlap between traversals within a given dataset. This facilitates the definition of novel dataset splits by geospatial overlap requirements across traversals.
    \item We introduce a novel semi-supervised training regime for \ac{bev} feature grid encoders, leveraging geospatial relations between different poses and their respective feature grids using a contrastive loss function.
    \item We evaluate our approach on the Argoverse~2 dataset against a set of supervised baselines, showing quantitative gains between 13\% and 42\%. Qualitatively, we demonstrate alignment of the latent space both across traversals and with target \ac{hd} maps.
\end{itemize}

%% file: sections/figures/overlap.tex
\vspace{2mm}
\begin{figure}[ht]
    \centering
    \adjustbox{trim=0cm 0cm 0cm 0cm, width=\columnwidth, clip}{
        \includegraphics[width=\columnwidth]{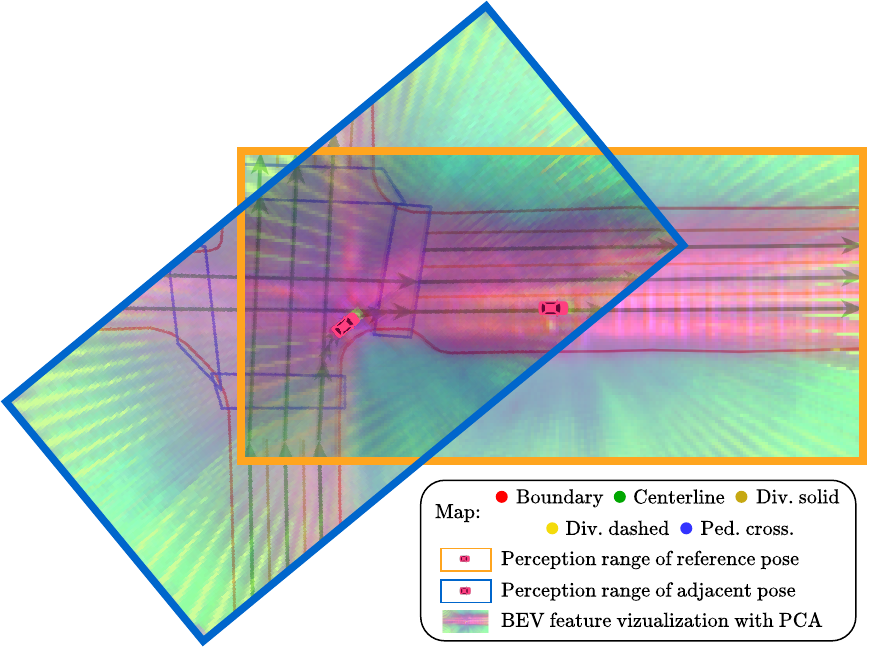}
    }
    \caption{Overlapping BEV feature grids visualized using PCA.
    In the background, we depict ground truth map labels with road boundaries, lane dividers, centerlines and pedestrian crossing. Both the geospatial consistency between overlapping BEV feature spaces as well as their alignment with the target road layout can be recognized.}
    \label{fig:overlap}
\end{figure}

%% file: sections/02_related_work.tex
\section{Related work}
\label{sec:related_work}

In the following section, we cover relevant prior work on semi-supervised learning and online \ac{hd} map construction.

\subsection{Semi-supervised Learning}

In the semi-supervised learning paradigm, a large unlabeled dataset and a small labeled dataset are combined to train a model, whereas in \ac{ssl}, pseudo-labeling is often used to generate labels from the data itself, and therefore, no real labels are needed. In general, two approaches are examined in semi-supervised learning: multi-stage training and one-stage training. In the multi-stage setup, the model is first trained unsupervised or self-supervised to learn generic features and then finetuned supervised for the actual downstream task~\autocite{simclr_chen_2020, caron_dino}. In contrast, in one-staged training, labeled and unlabeled data are mixed within a batch and processed with different data flows, resulting in different loss terms for the labeled and unlabeled data~\autocite{Tang_semi_sup_proposal_learning, Xu_semi_sup_soft_teacher, Chen_semi_sup_label_matching}.

While semi-supervised learning have had a strong influence on image classification and object detection in previous years, recent research has extended these concepts to the domain of autonomous driving. Several \ac{ssl} techniques, such as teacher-student architectures~\cite{Zhu_mean_teacher_sem_bev_seg}, feature distillation~\cite{Sirko-Galouchenko_occfeat} and contrastive learning~\autocite{leng_bevcon} have been leveraged to improve the performance of semantic \ac{bev} segmentation or 3D object detection in the autonomous driving domain.

In the online HD map construction domain, only two approaches leverage semi-supervised learning. PseudoMapTrainer~\cite{Lowens_pseudo_map_trainer} utilizes sensor data to generate a coherent mesh grid of Gaussian surfels. By rendering these surfels into a \ac{bev} segmentation, pseudo-labels are generated and used to pre-train the model in a semi-supervised manner. In contrast, Lilja et al.~\cite{Lilja_semi_supervised} employ a semi-supervised framework built on a teacher-student architecture. To improve the soft labels of the teacher network, a temporal pseudo-label fusion is introduced, which aggregates these labels over time.
Both methods leverage pseudo-labels generated from sensor data in a semi-supervised setup to boost model performance. In contrast, our approach exploits geospatial consistency to learn meaningful \ac{bev} features by training the encoder with an additional self-supervised objective, thereby improving performance on the online \ac{hd} map construction task. Another key distinction is that we utilize a vector-based method that directly predicts map elements as polylines, whereas PseudoMapTrainer~\cite{Lowens_pseudo_map_trainer} and Lilja et al.~\cite{Lilja_semi_supervised} employ a semantic segmentation-based approach.

\subsection{Online HD Map Construction}

HDMapNet~\cite{Li_hdmapnet} first proposed a multi-stage pipeline to predict a set of vectorized map elements by transforming camera images into a \ac{bev} grid and utilizing semantic segmentation with a heuristic postprocessing. Subsequent studies, including VectorMapNet~\cite{liu_vectormapnet} and MapTR~\cite{liao_MapTR}, utilize the transformer-style architecture DETR~\cite{carion_detr} as a decoder to directly predict map elements as polylines. Whereas VectorMapNet~\cite{liu_vectormapnet} utilizes an autoregressive approach to predict each instance, MapTR~\cite{liao_MapTR} introduces a hierarchical query design to predict all polylines in a single step. Based on this hierarchical query design, MapTRv2~\cite{liao2024maptrv2}, MapQR~\cite{liu_mapqr} and DAMap~\cite{dong_damap} extend this approach with auxiliary tasks, advanced query design, or loss function adaptations.

While these single-shot approaches utilize only current sensor information, recent studies attempt to boost performance by leveraging information from previous predictions in the form of short-term memory. HDMapNet~\cite{Li_hdmapnet} introduces short-term memory by fusing \ac{bev} features with max-pooling, whereas StreamMapNet~\cite{Yuan_2024_streammapnet} goes one step further by establishing a memory buffer with a \ac{gru} and storing \ac{bev} features and decoder queries from the previous predictions. By establishing a tracking mechanism for the predicted map elements and \ac{bev} features, MapTracker~\cite{chen_maptracker_2025} significantly improves the map construction performance and achieves state-of-the-art performance. 

To address the limitations of single-shot and short-term memory approaches, global map priors are used to enhance the quality of predictions, particularly in long-range scenarios. SMERF~\cite{luo_smerf}, SDTagNet~\cite{immel_SDTagNet2025} and M3TR~\cite{immel_m3tr} utilize \ac{sd} or outdated \ac{hd} maps to enhance recent sensor information. In contrast, HRMapNet~\cite{zhang2024hrmapnet}
exploits multi-traversals by storing \ac{bev} features or rasterized polylines into a global map. RTMap~\cite{du2025rtmap} extends the leverage of multi-traversals for online \ac{hd} map construction by adding uncertainties to local predictions.
Our objective is to investigate \ac{ssl} approaches that exploit geospatial consistency from multi-traversals to organize the \ac{bev} feature space in a meaningful way. To this end, we develop a semi-supervised framework based on a single-shot architecture that does not rely on previous predictions.

%% file: sections/03_methodology.tex
\input{sections/figures/overview}

\section{Methodology}
\label{sec:methodology}

This section describes the creation of a multi-traversal dataset split and our geospatial contrastive learning strategy to utilize geospatial consistency in a semi-supervised learning pipeline.
% The following \cref{subsec:geospatial} describes the utilization of multi-traversals for geospatial consistency (shown in \cref{fig:overlap}) and the creation of suitable dataset splits. Subsequently, our geospatial contrastive learning strategy is explained in \cref{subsec:spatial_aware_contrastive}, while \cref{subsec:semisup} describes our semi-supervised learning pipeline, as depicted in \cref{fig:semisupervised_pipeline_overview}.

\subsection{Geospatial Multi-traversal Split}
\label{subsec:geospatial}

As outlined in \cref{sec:related_work}, recent research on \ac{hd} map construction has focused on leveraging temporal or spatial consistency to improve model predictions. However, employing spatial consistency necessitates storing previous predictions at specific locations, resulting in vastly increased memory consumption and model complexity.

\noindent
\textbf{Single- and Multi-traversal Classification:}
To directly utilize spatial consistency across traversals, we propose a method for examining the geospatial overlap between traversals within a dataset.
This facilitates creating single- and multi-traversal splits based on these overlaps. Initially, we transform all poses into a global reference frame and divide the scenario logs into multiple large areas, \eg according to the various cities. Based on these divisions, we classify each log into single- and multi-traversals using the perception range of the vehicle. Consequently, we calculate a bounding box for each pose based on the vehicle heading denoting the driving direction, with $\pm x$~meters in the lateral and $\pm y$~meters in the longitudinal directions with respect to the vehicle's pose, and merge all the bounding boxes of each traversal into a single polygon. A traversal is classified as a multi-traversal if the corresponding polygon intersects at least one polygon from another traversal. If no other traversals intersects, the traversal is classified as a single-traversal. One exception is made: if only two drives intersect each other, both are classified as single-traversals. This exception is made since such that traversals provide only limited additional variety to effectively train for geospatial consistency and to enable an adequate size of the single-traversal split.

\noindent
\textbf{Spatial Graph Creation:}
To efficiently manage these overlaps, we construct a spatial graph $G=(V,E)$. Each vertex $v \in V$ represents a vehicle pose, and an edge $e_{ij} \in E$ is drawn between two poses if they satisfy the spatial intersection criterion. Specifically, we calculate the \ac{iou} between the perception grids of $P_i$ and $P_j$. Edges are maintained only for pairs in which the overlap falls within a predefined range [$\text{IoU}_\text{min}$, $\text{IoU}_\text{max}$]. Constraining the minimum and maximum values of the \ac{iou} ensures that the overlapping areas are sufficiently related but not identical.

\input{sections/figures/multidrive}
\subsection{Geospatial Contrastive Learning}
\label{subsec:spatial_aware_contrastive}

To leverage geospatial consistency in a multi-traversal setup, we propose a \ac{ssl} approach based on the contrastive learning framework SimCLR \cite{simclr_chen_2020}. Unlike traditional contrastive learning methods that rely on image augmentations, our approach exploits the inherent spatial correspondences that emerge when the same geospatial area is traversed multiple times. Specifically, we treat overlapping poses as natural augmentations and call them \textit{reference-adjacent} pairs, where the \textit{reference} denotes the reference pose and \textit{adjacent} denotes another pose with intersecting perception range being an adjacent in the spatial graph.

\noindent
\textbf{Positive and Negative Sample Definition:} 
Contrastive learning aims to organize the feature space by pulling similar data points together while pushing dissimilar points apart. This is achieved through positive and negative pair definitions relative to an anchor~\cite{simclr_chen_2020}. In our approach, the \ac{bev} grids $B_{\text{SSL,R}}$ and $B_{\text{SSL,A}}$ are predicted for the reference and adjacent poses based on the corresponding camera data. To employ geospatial consistency, we transform these \ac{bev} grids with respect to the vehicle's position and heading into a global coordinate system and utilize the \ac{bev} cells from these transformed grids to define positive and negative pairs relative to an anchor \ac{bev} cell. 
A \ac{bev} cell $c_p$ with feature vector $\mathbf{f}_p$ is considered a positive sample with respect to an anchor \ac{bev} cell $c_a$ with feature vector $\mathbf{f}_a$ if both cells represent the same geospatial location and therefore should exhibit similar features. Conversely, a \ac{bev} cell $c_n$ with feature vector $\mathbf{f}_n$ is considered a negative sample if the cells do not share spatial correspondence, resulting in dissimilar features.

% \todo[inline]{JP: Hier wird direkt von den schön motivierten **Posen**-Paaren zu BEV grids und v.a. auch negativen Samples gesprungen. Dazwischen fehlt noch a) was negativ bedeutet und b) wie genau jetzt Zellen im Bezug zur Pose stehen. Wieso wurde der Absatz hiernach auskommentiert? Alternativ taugt der Absatz mit der Sampling Strategy hierzwischen.}

\noindent
\textbf{Sampling Strategy:} 
Starting from two intersecting poses, the reference pose $R$ and the adjacent pose $A$, we categorize each of the \ac{bev} cells $c$ in the \ac{bev} grids $B_{\text{SSL,R}}$ and $B_{\text{SSL,A}}$ as inliers or outliers with respect to the overlap area.
For positive pairs, we randomly sample \ac{bev} cells $c$ from our \textit{reference} \ac{bev} grid $B_{\text{SSL,R}}$ that lie within the overlap area and serve as anchor points. For each of these anchor points, we perform a nearest-neighbor search in the \textit{adjacent} \ac{bev} grid to identify the corresponding positive cell at the same geospatial location. 
In contrast, the negative pairs are randomly sampled from both the reference and adjcent grid, explicitly excluding the anchor and positive cells to ensure the representation of distinct geospatial locations.
% \todo[inline]{Wieso wurden hier keine Zellen von BEV grids ohne Überlapp genommen? Das macht die Definition von "positiven" Posen egt hinfällig, da es keine "negativen" Posen, sondern nur Zellen gibt?}

\noindent
\textbf{Loss Function:} 
Based on the sampling of positive and negative cell pairs, we employ the InfoNCE loss \cite{infonce_oord_2019} to train the network. Analogous to SimCLR \cite{simclr_chen_2020}, we utilize a projection head $h$ and transform each \ac{bev} cell feature $\mathbf{f}$ into another feature space $\mathbf{z} \in \mathcal{Z}$ and therefore decouple the learning domain from the application domain. Given an anchor cell embedding $\mathbf{z}_i$ and its corresponding positive cell embedding $\mathbf{z}_i^+$, along with a set of negative embeddings $\{\mathbf{z}_k^-\}_{k=1}^{K}$, the geospatial contrastive loss $\mathcal{L}_{\text{GCLR}}$ is formulated as
\begin{equation}
\mathcal{L}_{\text{GCLR}} = -\log \frac{\exp\!\left( \frac{\text{sim}(\mathbf{z}_i, \mathbf{z}_i^+)}{\tau}\right)}{\exp\!\left(\frac{\text{sim}(\mathbf{z}_i, \mathbf{z}_i^+)}{\tau}\right) + \sum_{k=1}^{K}\exp\!\left(\frac{\text{sim}(\mathbf{z}_i, \mathbf{z}_k^-)}{\tau}\right)},
\end{equation}

where $\text{sim}(\cdot, \cdot)$ denotes the cosine similarity and $\tau$ is a temperature parameter. This objective encourages the network to produce similar embeddings for \ac{bev} cells corresponding to the same geospatial location across different traversals while pushing apart the embeddings of cells from different locations.
An overview of the geospatial contrastive learning approach for a positive cell pair is shown with orange and blue colors for the \textit{reference} and \textit{adjacent}, respectively, in \cref{fig:semisupervised_pipeline_overview}.

\input{sections/tables/comparison_table}

\subsection{Semi-supervised Training Regime}
\label{subsec:semisup}

This work aims to exploit \ac{ssl} to boost the performance of the online \ac{hd} map construction task. We propose to combine supervised and self-supervised methods in online \ac{hd} map construction under the assumption that we have access to two distinct datasets: 
a small, labeled dataset containing camera images with corresponding ground truth \ac{hd} map labels, which is expensive to annotate, as well as an easily scalable unlabeled dataset containing only camera images and vehicle poses, without any assumption on geographic or scene-level overlap with the labeled dataset.
\cref{fig:semisupervised_pipeline_overview} shows an overview of the proposed semi-supervised learning pipeline, where three distinct data flows are visualized: the supervised data flow (pink) and two self-supervised data flows for anchor and positive samples (orange and blue).

\noindent
\textbf{Architecture and Batch Composition:} 
From an architectural perspective, we adopt the single-shot \ac{hd} map construction model MapTRv2~\cite{liao2024maptrv2}, which follows an encoder-decoder structure. For each training batch, we sample $n$ poses with their corresponding camera images and ground truth map labels from the labeled dataset and fill-up the batch with samples from the unlabeled dataset. Since our contrastive approach requires \textit{reference-adjacent} pose pairs, where each reference pose $R$ is matched with a geospatially intersecting pose $A$, the number of unlabeled samples is constrained to be divisible by two. Consequently, each batch contains a total of $n + 2m$ samples, where $n$ denotes the number of supervised samples and $2m$ denotes the number of self-supervised samples organized as $m$ \textit{reference-adjacent} pairs.

\noindent
\textbf{Feature Extraction and Lifting:}
All samples are passed through a MapTRv2~\cite{liao2024maptrv2} map encoder, which consists of a ResNet-50 \cite{he_resnet} backbone for extracting image features and a lifting module for transforming these features into \ac{bev} representation. As a result, we obtain \ac{bev} grids for each sample: Specifically, $n$ \ac{bev} grids $\mathcal{B}_\text{sup}$ for the supervised samples, $m$ \ac{bev} grids $\mathcal{B}_\text{SSL,R}$ for the reference poses, and $m$ \ac{bev} grids $\mathcal{B}_\text{SSL,A}$ for the adjacent poses from the unlabeled dataset.

\noindent
\textbf{Supervised Learning Branch:}
From this point, the data flow diverges between labeled and unlabeled data. The labeled data, represented by \ac{bev} grids $\mathcal{B}_\text{sup}$, flows directly into a map decoder based on MapTRv2~\cite{liao2024maptrv2}. This decoder utilizes a transformer-based architecture to predict map elements in the form of polylines with corresponding class labels. The supervised loss $\mathcal{L}_\text{sup}$, strictly following MapTRv2~\cite{liao2024maptrv2}, is computed based on the predicted polylines and classes and backpropagated through both the map decoder and map encoder modules. For each batch, $\mathcal{L}_\text{sup}$ is calculated as the sum over all $n$ labeled samples.

\noindent
\textbf{Self-supervised Learning Branch:}
For the unlabeled data, the generated \ac{bev} grids $\mathcal{B}_\text{SSL,R}$ and $\mathcal{B}_\text{SSL,A}$ from the reference and adjacent pose are used to sample the positive and negative cell pairs and compute the geospatial contrastive loss $\mathcal{L}_{\text{GCLR}}$, as described in Section~\ref{subsec:spatial_aware_contrastive}. Again, $\mathcal{L}_{\text{GCLR}}$ is computed as sum over all $m$ unlabeled pose pairs.

\noindent
\textbf{Combined Loss Function:}
The overall semi-supervised loss $\mathcal{L}_{\text{semi}}$ for each batch is determined by the weighted combination of the supervised loss $\mathcal{L}_\text{sup}$ and geospatial contrastive loss $\mathcal{L}_\text{GCLR}$, with weighting factors $\lambda_{\text{sup}}$ and $\lambda_{\text{GCLR}}$, respectively:

\begin{equation}
\mathcal{L}_{\text{semi}} = \lambda_{\text{sup}} \mathcal{L}_{\text{sup}} + \lambda_{\text{GCLR}} \mathcal{L}_{\text{GCLR}}.
\label{eq:semi_supervised_loss}
\end{equation}

These weighting factors serve two purposes: they scale both loss terms into comparable ranges and allow us to control the relative influence of supervised and \ac{ssl} on the overall training objective. This formulation enables the model to benefit from both the precise supervisory signal from labeled data and the geospatial consistency constraints learned from unlabeled multi-traversal data.

%% file: sections/figures/overview.tex
\begin{figure*}[t]
    \centering
    \includegraphics[width=\textwidth]{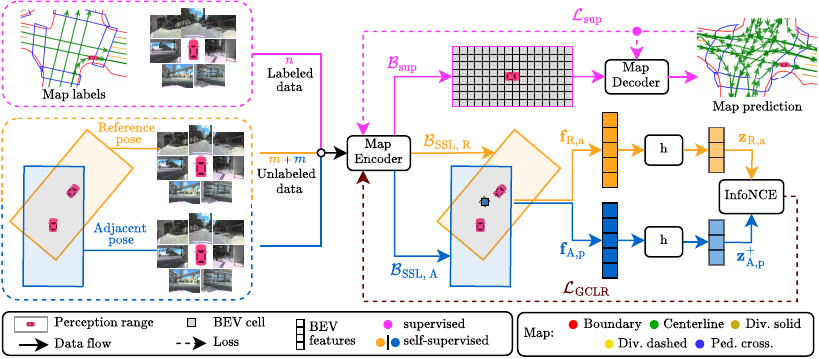} 
    \caption{Schematic overview of the semi-supervised learning pipeline. Data flows are shown for supervised (pink) and self-supervised (blue, orange) samples. $\mathcal{B}$ denotes \ac{bev} grids with cell features $\mathbf{f}$ and embeddings $\mathbf{z}$ produced by projection head $h$. Losses $\mathcal{L}_{\text{sup}}$ and $\mathcal{L}_{\text{GCLR}}$ correspond to supervised and self-supervised branches, respectively.}
    \label{fig:semisupervised_pipeline_overview}
\end{figure*}

%% file: sections/figures/multidrive.tex
\begin{figure*}[t]
\centering
\begin{tabular}{cc}
% \includesvg[width=0.476\textwidth]{images/multidrive_histogram.svg} &
% \includesvg[width=0.476\textwidth]{images/MIA.svg} \\
\includegraphics[width=0.476\textwidth]{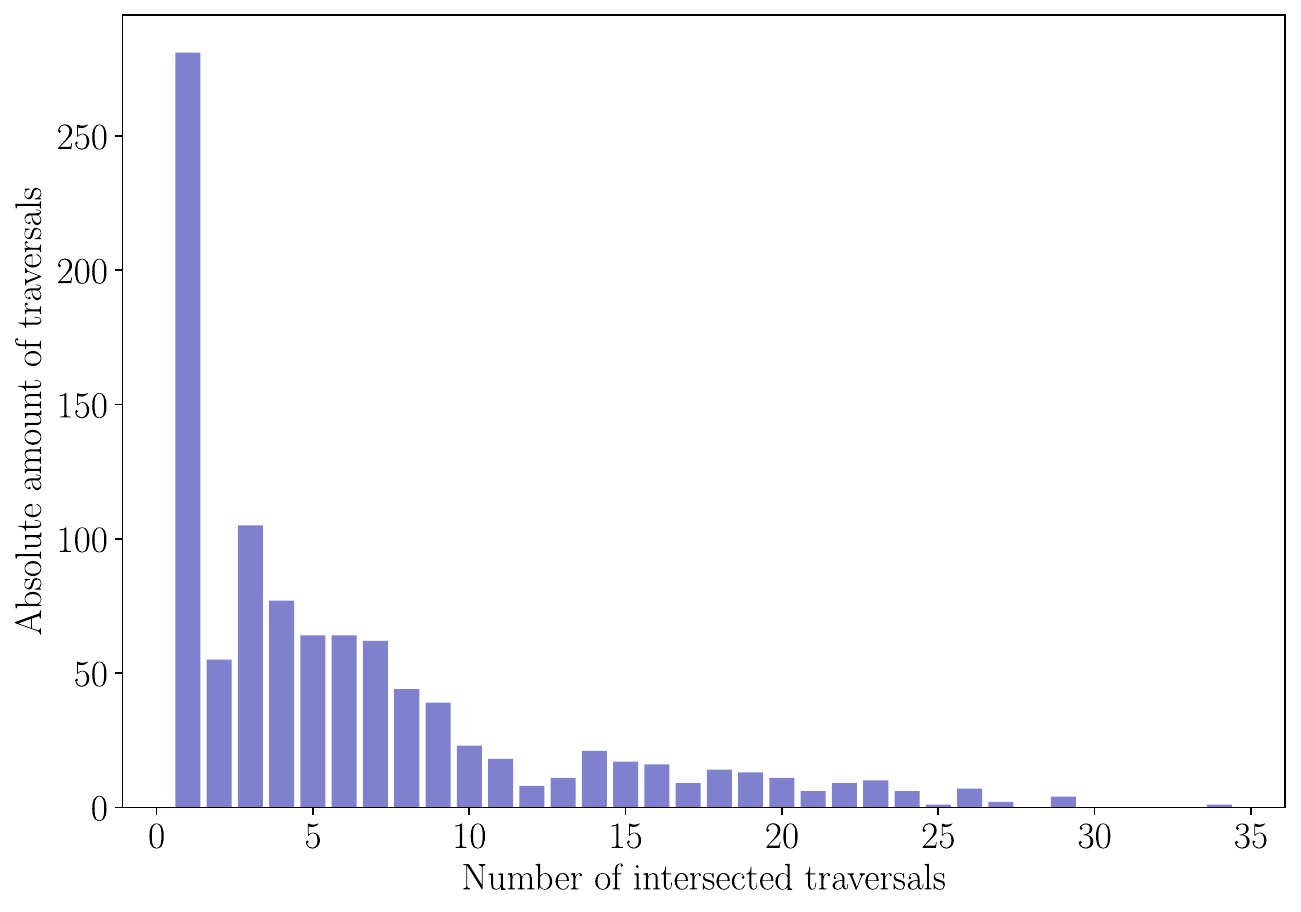} &
\includegraphics[width=0.476\textwidth]{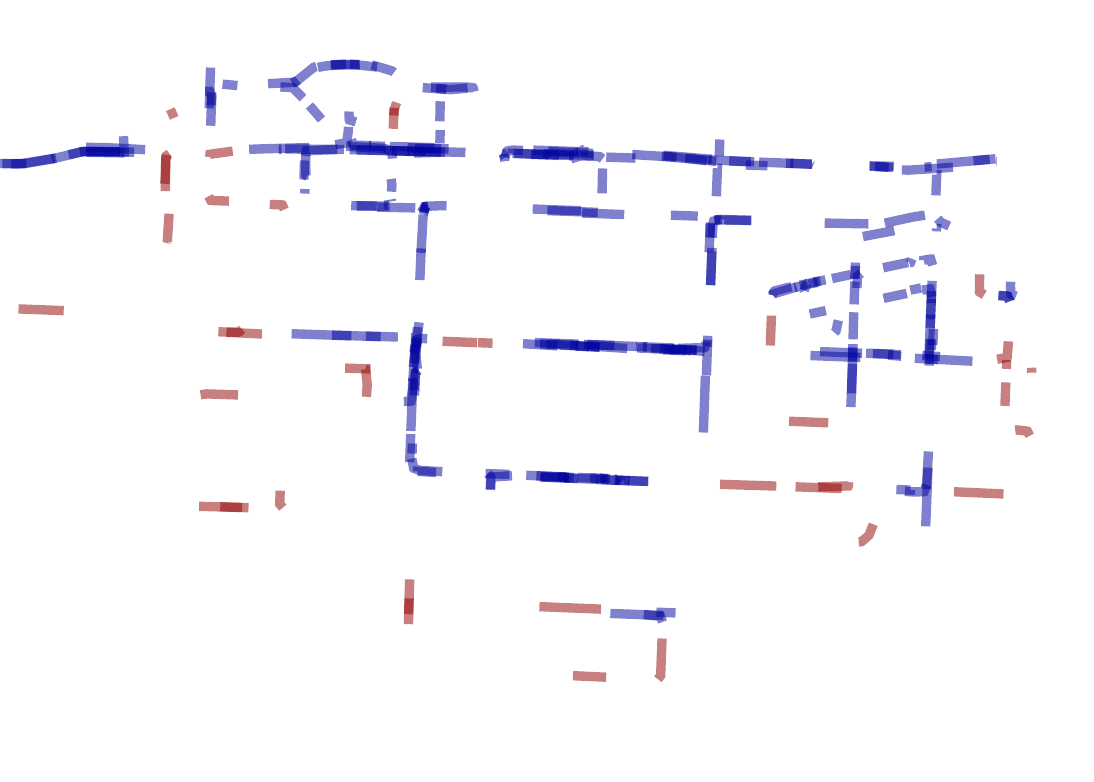} \\
\end{tabular}
\caption{Single- and multi-traversals within Argoverse 2: The histogram on the left shows the general distribution of intersecting drive logs. On the right, we visualize the geospatial distribution of single- (red) and multi-traversals (blue) within Miami.
}
\label{fig:multidrive}
\end{figure*}
% \todo[inline]{Figure 3: graue durchsichtige kacheln in den Hintergrund?}

%% file: sections/tables/comparison_table.tex
{ % new env for renewcommands
\renewcommand{\arraystretch}{1.15}

\begin{table*}[t]
    \centering
    % In IEEEtran, table captions must be placed above the table
    \caption{Comparison of the supervised and semi-supervised training strategies for the categories divider dashed~(dsh), solid~(solid), boundary~(bou), centerline~(cen) and pedestrian crossing~(ped). The first two columns denote the share of supervised training data as well as the use of self-supervised learning (SSL).}
    \label{tab:map}
    \begin{tabular}{c | c | c c c c c | c c c}
    \hline
    Sup. Data & SSL & AP$_{\text{dsh}}$ & AP$_{\text{sol}}$ & AP$_{\text{bou}}$ & AP$_{\text{cen}}$ & AP$_{\text{ped}}$ & mAP & $\Delta_\mathrm{abs}$ & $\Delta_\mathrm{rel}$ \\
    \hline
    
    % ===== 2.5% =====
    \multirow{2}{*}{2.5\%} & \ding{55} & 4.3 & 5.0 & 9.6 & 11.9 & 1.5 & 6.5 & \multirow{2}{*}{\color{Green}\bfseries +2.0} & \multirow{2}{*}{\color{Green} \bfseries +31\%}\\
    & \ding{51} & 5.2 & 6.7 & 12.2 & 17.0 & 1.6 & 8.5 \\
    \hline
    
    % ===== 5% =====
    \multirow{2}{*}{5\%} & \ding{55} & 10.3 & 9.5 & 20.5 & 19.1 & 7.3 & 13.3 & \multirow{2}{*}{\color{Green}\bfseries +5.6} & \multirow{2}{*}{\color{Green} \bfseries +42\%}\\
    & \ding{51} & 15.4 & 18.7 & 24.8 & 25.4 & 9.9 & 18.9 \\
    \hline
    
    % ===== 10% =====
    \multirow{2}{*}{10\%} & \ding{55} & 17.6 & 20.9 & 31.9 & 27.1 & 12.4 & 22.0 & \multirow{2}{*}{\color{Green}\bfseries +5.3} & \multirow{2}{*}{\color{Green} \bfseries +24\%}\\
    & \ding{51} & 20.8 & 30.5 & 34.5 & 32.4 & 18.2 & 27.3 \\
    \hline
    
    % ===== 20% =====
    \multirow{2}{*}{20\%} & \ding{55} & 27.2 & 32.1 & 38.9 & 34.7 & 22.3 & 31.0 & \multirow{2}{*}{\color{Green}\bfseries +3.9} & \multirow{2}{*}{\color{Green} \bfseries +13\%}\\
    & \ding{51} & 31.2 & 38.8 & 39.9 & 37.5 & 26.9 & 34.9 \\
    \hline

    % ===== 30% =====
    %\multirow{2}{*}{30\%} & \ding{55} & 34.3 & 39.8 & 41.7 & 38.7 & 28.7 & 36.6 & & \\
    %& \ding{51} & - & - & - & - & - & - & & \\
    30\% & \ding{55} & 34.3 & 39.8 & 41.7 & 38.7 & 28.7 & 36.6 & & \\
    \hline

    % ===== 40% =====
    %\multirow{2}{*}{40\%} & \ding{55} & 37.8 & 43.7 & 43.9 & 41.2 & 32.7 & 39.8 & & \\
    %& \ding{51} & - & - & - & - & - & - & & \\
    40\% & \ding{55} & 37.8 & 43.7 & 43.9 & 41.2 & 32.7 & 39.8 & & \\
    \hline
    \end{tabular}
\vspace{-2mm}
\end{table*}
} % end new env

%% file: sections/04_evaluation.tex
\input{sections/figures/rel_gain_plot}

\section{Evaluation}

\subsection{Experimental Setup}
We evaluate our approach on Argoverse~2~\cite{Argoverse2}, a well-established dataset for supervised online \ac{hd} map construction~\autocite{liao2024maptrv2,chen_maptracker_2025}.
However, its fit for semi-supervised training as described in~\cref{subsec:semisup} remains to be investigated.

\noindent
\textbf{Multi-Traversal Analysis:}
% In order for Argoverse~2 to qualify as a solid foundation for our semi-supervised experiments, we require a great number of multi-traversals present within the dataset. We employ our classification approach as described in~\cref{subsec:geospatial}, providing us with deep insight into the composition of the dataset as visualized both quantitatively and qualitatively in~\cref{fig:multidrive}. The qualitative visualization demonstrates the effectiveness of our classification step with majorly overlapping sections consistently being classified as multi-traversal. The histogram of~\cref{fig:multidrive} states the presence of a vast portion of traversals within the dataset to include multiple intersections with other traversals, making Argoverse 2 well suited for evaluating our training approaches.
In order for Argoverse~2 to qualify as a solid foundation for our semi-supervised experiments, we require a great number of multi-traversals present within the dataset. Therefore, we analyze the prevalence of multi-traversal sequences using our geospatial classification approach described in~\cref{subsec:geospatial}. The qualitative visualization demonstrates the effectiveness of our classification step with majorly overlapping sections consistently being classified as multi-traversal. The histogram of~\cref{fig:multidrive} states that the majority of logs contain multiple 
intersections with other traversals, confirming that Argoverse~2 is well suited for evaluating our training approach

\input{sections/figures/pca_comparison}

\noindent
\textbf{Dataset Splits:}
% Based on our single- and multi-traversal classification from in~\cref{subsec:geospatial}, we utilize all logs classified as multi-traversal as our self-supervised split, ignoring labels present for this portion in all experiments all together. Among those classified as single-traversal, we initially select logs at random to form a static validation set comprising 10~\% of the total number of poses in the dataset, always including entire logs. From the remaining single-traversals, we randomly shuffle and sequentially assign complete logs to our supervised subsets, creating splits that comprise 2.5~\%, 5~\%, 10~\%, and 20~\% of the total number of poses in the dataset. We define these static sets this way to ensure lower percentage supervised subsets to also always be a subset of all its higher percentage subsets, reducing selection bias between subsets. Higher percentage supervised subsets (30~\% and 40~\% splits) are defined to provide a trend of the supervised performance, include parts of the self-supervised split and are therefore not usable for self-supervised training.
Based on the classification from~\cref{subsec:geospatial}, all multi-traversal logs form the self-supervised split, with their labels discarded throughout all experiments. From the remaining single-traversal logs, 10~\% of total poses are reserved as a static validation set. The remaining logs are randomly shuffled and sequentially assigned to supervised subsets of 2.5~\%, 5~\%, 10~\%, and 20~\% of total poses, ensuring that smaller subsets are always contained within larger ones to reduce selection bias. Supervised subsets of 30~\% and 40~\% are additionally provided to indicate the trend of purely supervised performance, but overlap with the self-supervised split and are therefore excluded from semi-supervised training.

\noindent
\textbf{Hyperparameters:}
All hyperparameters are determined empirically via scheduled experiments, yielding $\lambda_{\text{sup}}=1.0$, $\lambda_{\text{self}}=0.1$, $\tau=0.25$, $n_\text{anchors}=1023$, and $n_\text{negatives}=1024$, where $n_\text{anchors}+1=1024$ is chosen as a power of two for memory efficiency. The projection head $h$ employs a hidden dimension of 256 and an output dimension of 64. The spatial graph is constructed with an IoU range of $[0.2,\, 0.8]$. All models are trained on 4$\times$A100 GPUs with a batch composition of 1 labeled and 4 unlabeled samples per GPU, constrained by available VRAM. Following~\cite{liao2024maptrv2}, all models are optimized using AdamW with a learning rate of $3.75\times10^{-4}$.

\subsection{Quantitative Results}
To quantify the benefit of our semi-supervised training regime, we employ MapTRv2~\cite{liao2024maptrv2} trained with its original purely supervised training regime as baseline. To level the playing field, we provide both models with equal access to labeled training data, varying the percentage based on the split at hand. \cref{tab:map} presents the increase in performance by utilizing self-supervised training (SSL) compared to the baseline. The benefit of the proposed method is also visualized in \cref{fig:trend}.

Using the proposed self-supervised contrastive learning with additional unlabeled training data significantly improves the online \ac{hd} map construction performance across all conducted experiments, yielding relative gains between 13~\% and 42~\%.
As apparent in \cref{fig:trend}, the benefit is particularly distinct for small amounts of labeled data, where it is almost as good as doubling the amount of labeled training data.

\subsection{Qualitative Results}
We visualize the \ac{bev} feature grid using \ac{pca} to qualitatively assess its spatial and semantic structure, as coherent and well-separated feature organization is indicative of effective representation learning and typically correlates with improved downstream performance~\cite{simeoni2025dinov3}. The \ac{pca} is calculated across all scenes of the validation set to provide consistent results.

Two example scenes can be observed in~\cref{fig:pcl_comparison}. While the feature separation of both the baseline and our semi-supervised approach appears cleaner in the less complex upper scene, ours delivers stronger contrasts in both scenes, especially around road borders. Additionally, a stronger separation of the ego lane can be observed.

Interestingly, we also observe an unexpected feature cluster in the baseline output in the top right of its grid, that always appear at identical grid coordinates, a curiosity that is effectively eliminated with out approach, since it contradicts geospatial consistency.

%\subsection{Ablation study}

%\begin{itemize}
%    \item different weightings
%\end{itemize}

%% file: sections/figures/rel_gain_plot.tex
\begin{figure}[t]
\vspace{2mm}
\centering
\begin{tikzpicture}

% Define futuristic colors
\definecolor{neonblue}{RGB}{0,180,255}
\definecolor{neonmagenta}{RGB}{255,0,140}
\definecolor{neongreen}{RGB}{0,200,120}

% -------- Left axis (mAP) --------
\begin{axis}[
    width=0.95\columnwidth,
    height=6cm,
    xlabel={\textbf{Share of Supervised Data (\%)}},
    ylabel={\textbf{mAP (\%)}},
    xmin=0, xmax=42,
    ymin=0,
    ymax=46,
    xtick={2.5,5,10,20,30,40},
    grid=both,
    major grid style={line width=0.4pt, draw=gray!25},
    minor grid style={line width=0.2pt, draw=gray!15},
    axis line style={black!80, thick},
    tick style={black!80},
    tick label style={font=\small},
    label style={font=\bfseries},
    clip=false,
    legend style={
        at={(0.51,0.4)},
        anchor=north west,
        legend columns=1,
        draw=gray!30,
        fill=white,
        rounded corners=3pt,
        font=\small,
        column sep=0.1cm
    },
]

% --- Supervised ---
\addplot[
    smooth,
    ultra thick,
    color=neonmagenta,
    mark=*,
    mark options={scale=1.2, fill=neonmagenta}
] coordinates {
    (2.5, 6.5)
    (5,   13.3)
    (10,  22.0)
    (20,  31.0)
    (30,  36.6)
    (40,  39.8)
};
\addlegendentry{Supervised}

% --- Semi-supervised ---
\addplot[
    smooth,
    ultra thick,
    color=neonblue,
    mark=square*,
    mark options={scale=1.1, fill=neonblue}
] coordinates {
    (2.5, 8.5)
    (5,   18.9)
    (10,  27.3)
    (20,  34.9)
};
\addlegendentry{Semi-supervised}

% --- Relative Gain (Right Axis) ---
\addplot[
    smooth,
    very thick,
    dashed,
    color=neongreen,
    mark=o,
    mark options={scale=1.2, fill=neongreen}
] coordinates {
    (2.5, 30.8)
    (5,   42.1)
    (10,  24.1)
    (20,  12.6)
};
\addlegendentry{Relative Gain}

\end{axis}

% -------- Right axis (Relative Gain, invisible axis lines) --------
% \begin{axis}[
%     width=0.97\columnwidth,
%     height=6cm,
%     xmin=0, xmax=22,
%     ymin=0, ymax=50,
%     xtick=\empty,
%     ytick=\empty,
%     axis y line*=right,
%     axis x line=none,
%     ylabel={\textbf{Relative Gain (\%)}},
%     axis line style={black!0}, % hide right axis line
%     tick style={black!0} % hide ticks
% ]
% \end{axis}

\begin{axis}[
    width=0.95\columnwidth,
    height=6cm,
    xmin=0, xmax=42,
    ymin=0, ymax=46,
    xtick=\empty,
    ytick={0,20,40},
    yticklabels={0,20,40},
    axis y line*=right,
    axis x line=none,
    ylabel={\textbf{Relative Gain (\%)}},
    axis line style={black!80, thick},
    tick style={black!80},
    tick label style={font=\small},
]
\end{axis}

\end{tikzpicture}
\caption{Performance scaling and relative gains across increasing share of supervised data.}
\label{fig:trend}
\end{figure}
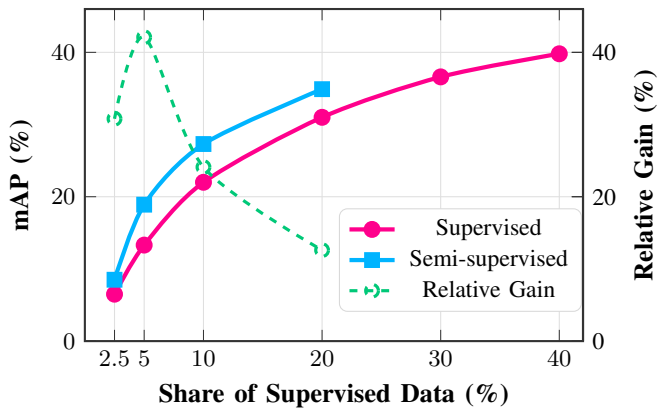

%% file: sections/figures/pca_comparison.tex
\begin{figure*}[t]
    \centering
    \includegraphics[width=\textwidth]{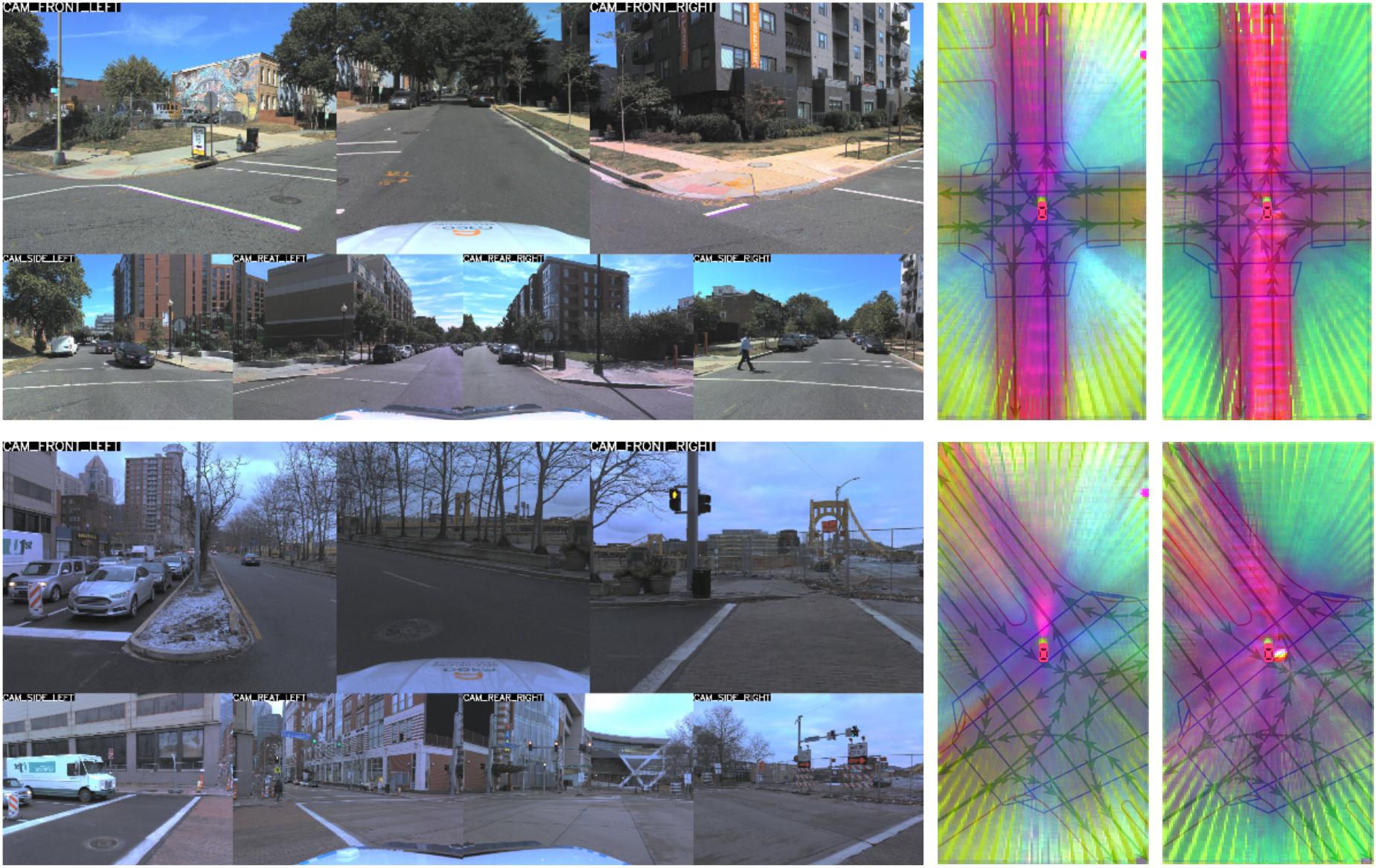}
    \caption{Qualitative visualization of PCA of the supervised baseline (middle) and our semi-supervised approach (right) with ground-truth labels in the background for reference. Input surround view on the left, 20~\% supervised split.}
    \label{fig:pcl_comparison}
\end{figure*}

%% file: sections/05_conclusion.tex
\section{Conclusion}
We presented an effective approach to boost online vectorized HD map construction performance using geospatial contrastive learning with unlabeled data.
It leverages the geospatial consistency across traversals in \ac{bev} feature space.
By implementing an effective single- and multi-traversal classification for autonomous driving datasets, we not only pave ground work for our own semi-supervised training regime, but also provided statistical insights into the Argoverse 2 dataset in particular.
Our semi-supervised training regime outperforms the supervised baseline across the board, demonstrating the effectiveness of our approach both quantitatively in terms of mAP and qualitatively with improved separation in \ac{bev} feature space.

A major prerequisite for our approach is the availability of highly accurate (relative) localization, which is also a crucial requirement for purely supervised models that use a ground truth map~\cite{BlumbergMerkertFehler2025_1000190486}.
While this feature is unfortunately missing in some of the largest autonomous driving datasets that would allow scaling our idea even further~\cite{nvidia_physicalai_autonomous_vehicles_2025}, our contrastive loss formulation could also be used to refine relative poses, mitigating this bottleneck at least partially.
Further work can also extend our approach by incorporating \ac{ssl} methods into the transformer-decoder of the model.
This could propagate the benefits beyond the intermediate \ac{bev} representation, boosting the final map prediction performance of the model. 

%% file: sections/06_acknowledgement.tex
\section*{Acknowledgements}

The authors gratefully acknowledge the computing time provided on the high-performance computer HoreKa by the National High-Performance Computing Center at KIT (NHR@KIT). This center is jointly supported by the Federal Ministry of Education and Research and the Ministry of Science, Research and the Arts of Baden-Württemberg, as part of the National High-Performance Computing (NHR) joint funding program (https://www.nhr-verein.de/en/our-partners). HoreKa is partly funded by the German Research Foundation (DFG).